\documentclass[11pt]{article}
\usepackage{iclr2027_conference,times}
\iclrfinalcopy  

\usepackage{graphicx}
\usepackage{amsmath}
\usepackage{amssymb}
\usepackage{algorithm}
\usepackage{algpseudocode}
\usepackage{booktabs}
\usepackage{multirow}
\usepackage{array}
\usepackage{xcolor}
\usepackage{colortbl}
\usepackage{makecell}
\usepackage{enumitem}
\usepackage{float}
\usepackage{stfloats}
\usepackage{placeins}
\usepackage{tikz}
\usetikzlibrary{arrows.meta,positioning,fit,backgrounds,calc}
\usepackage[most]{tcolorbox}
\newtcolorbox{keybox}{colback=green!6,colframe=green!45!black,boxrule=0.5pt,arc=2pt,
  left=5pt,right=5pt,top=3pt,bottom=3pt,before skip=3pt,after skip=3pt,fontupper=\small\itshape}
\usepackage[pagebackref,breaklinks,colorlinks]{hyperref}
\usepackage{url}
\usepackage{etoc}   

\newcommand{\morph}{\textbf{MIST}}
\newcommand{\best}[1]{{\bfseries\boldmath #1}}

\newcommand{\heavytext}[1]{\leavevmode\rlap{\kern0.22pt #1}#1}

\makeatletter
\renewcommand{\@opargbegintheorem}[3]{%
  \trivlist\item[]\noindent{\large\bfseries\heavytext{#1\ #2\ (#3):}}\enspace\normalfont}
\makeatother
\newcommand{\insightbox}[2]{%
  \par\noindent
  \begingroup
  \setlength{\fboxsep}{7pt}%
  \fcolorbox{blue!55!black}{blue!4}{%
    \parbox{\dimexpr\linewidth-2\fboxsep-2\fboxrule\relax}{%
      \textbf{\textcolor{blue!55!black}{#1}}\quad #2}}%
  \endgroup\par}

\begin{document}

\title{MIST: Multi-Scale Spatial Context for Gene Expression Prediction from Histology}

\author{%
Koushik Howlader$^{1}$\thanks{Corresponding author: \texttt{howlader@iastate.edu}.}
\quad Tirtho Roy$^{1}$
\quad Md Tauhidul Islam$^{2}$
\quad Wei Le$^{1}$ \\[3pt]
$^{1}$Iowa State University, Iowa, USA \qquad
$^{2}$Stanford University, CA, USA \\
}
\maketitle
\lhead{Preprint. Under review.}

\begin{abstract}
Predicting spatial gene expression from standard H\&E images could extend molecular analysis to tissues without spatial transcriptomic measurements. A standard Transformer applies self-attention over spots but ignores where each spot sits in the tissue, whereas a spot's molecular state depends on tissue context at multiple spatial scales. Building on this insight, we introduce \morph{}, a \textbf{M}ult\textbf{I}-context \textbf{S}patial \textbf{T}ransformer that augments the self-attention backbone with two context pathways applied in every layer: a coordinate-invariant \emph{local} pathway that attends over spatial \(k\)-nearest-neighbor spots, and a \emph{slide} pathway that broadcasts a whole-slide summary to every spot; the model predicts all spots in a single forward pass. Because coordinates enter only through pairwise distances, the local pathway is invariant to translations, rotations, and reflections when image features are fixed. In intra-cohort evaluation across six human organ cohorts from HEST-1k, \morph{} improves mean Pearson correlation over the strongest evaluated baseline by 9.7\%, 11.6\%, and 11.4\% for 10-, 50-, and 100-gene panels, respectively, with the lowest average MSE and MAE. \morph{} further improves leave-one-organ-out transfer by 6.0\% over the next strongest baseline, attains the best expression-derived domain recovery (highest mean ARI and NMI at every clustering resolution, and higher NMI than the closest baseline on 84\% of held-out slides), and reproduces spatial expression patterns of selected cancer-marker genes. Together, these results show that adding local and slide context to a vanilla Transformer backbone is an effective approach to spatial gene expression prediction from histology. Our code is available at \url{https://anonymous.4open.science/r/MIST-F660/}.
\end{abstract}

\etocdepthtag.toc{mainbody}
\section{Introduction}

Spatial transcriptomics (ST) maps gene expression to tissue locations,
revealing spatial heterogeneity and tissue organization
\cite{staahl2016visualization}. However, its cost and experimental
requirements limit widespread use. Routine H\&E images contain
morphological signals associated with molecular state, motivating the
prediction of spatial gene expression without matched ST measurements
\cite{chen2024stimage,jaume2024hest}.

A standard Transformer over tissue spots uses self-attention, allowing
each spot to attend to all others, but does not explicitly encode where spots
lie in the tissue. Existing H\&E-to-ST methods use patch-level regression
\cite{he2020stnet}, graph and transformer modeling
\cite{zeng2022hist2st}, cross-modal retrieval
\cite{xie2023bleep}, or conditional diffusion
\cite{zhu2025diffusion}. Yet a spot's molecular state may depend on context
beyond self-attention alone, particularly its immediate spatial neighborhood
and the overall tissue state. Neither is explicitly modeled by a plain
self-attention Transformer. We therefore ask:

\begin{center}
\textit{\textbf{Does augmenting self-attention with local spatial
and whole-slide context improve gene expression prediction from histology?}}
\end{center}

\begin{figure*}[t]
    \centering
    \includegraphics[width=1.00\linewidth]{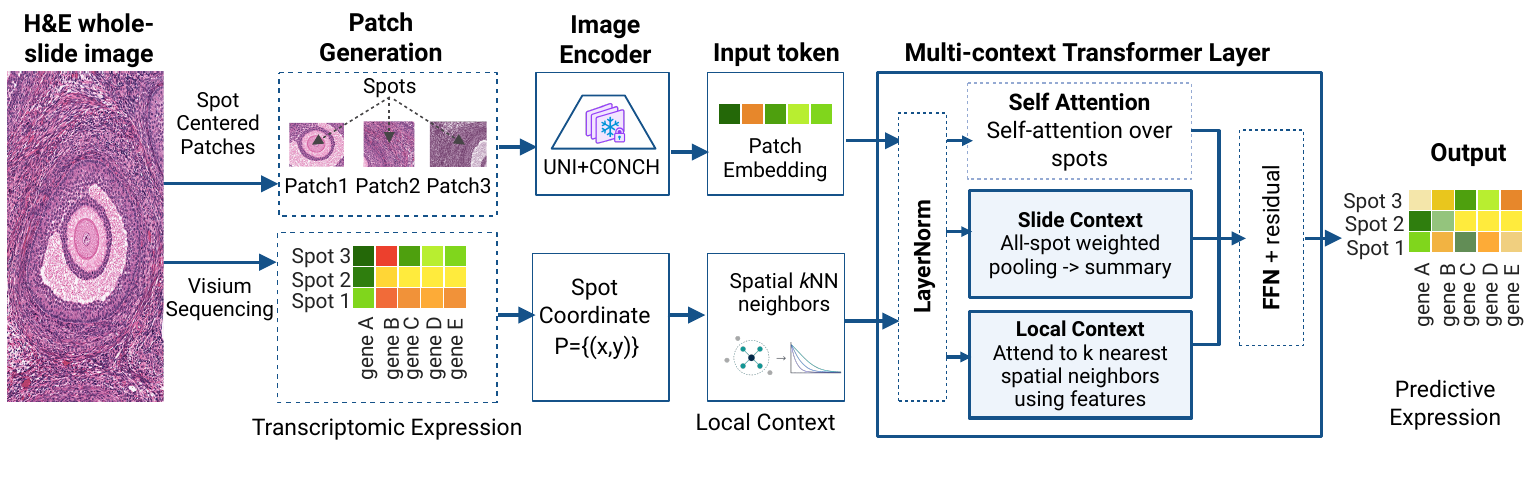}
    \caption{
        \textbf{\morph{} overview.}
        \textbf{(a)} Frozen UNI and/or CONCH embeddings represent
        spot-centered H\&E patches, while spot coordinates define
        spatial $k$NN neighborhoods.
        \textbf{(b)} Each Transformer layer augments the
        self-attention backbone with two pathways: local
        spatial-neighbor attention and an attention-pooled
        slide summary broadcast to all spots. The final spot
        representations are mapped to predicted gene expression.
    }
    \label{fig:arch}
\end{figure*}

We introduce \morph{}, which augments a standard self-attention
Transformer with two pathways applied in every layer
(Fig.~\ref{fig:arch}). Starting from frozen UNI and/or CONCH embeddings
\cite{chen2024uni,lu2024conch}, the \emph{local} pathway attends over each
spot's spatial $k$-nearest neighbors, while the \emph{slide} pathway forms
an attention-weighted whole-slide summary that is broadcast to every spot.
Their outputs are added to the self-attention backbone and refined through
residual feed-forward updates. The model predicts expression for all spots
in a single forward pass.

Coordinates are used only to construct the spatial $k$NN graph from
pairwise distances. Because these distances are unchanged by translations,
rotations, and reflections, the local spatial pathway is coordinate-frame
invariant for fixed image features, deterministic evaluation, and consistent
handling of $k$NN distance ties~\cite{satorras2021egnn}.

We evaluate \morph{} on six human organ cohorts from
HEST-1k~\cite{jaume2024hest} under intra-cohort, pooled multi-organ,
and leave-one-organ-out settings. In intra-cohort evaluation,
\morph{} improves mean Pearson correlation over the strongest evaluated
baseline by 9.7\%, 11.6\%, and 11.4\% for 10-, 50-, and 100-gene panels,
respectively, while achieving the lowest average MSE and MAE. In
leave-one-organ-out evaluation, it improves mean correlation by 6.0\%
over BLEEP~\cite{xie2023bleep}. \morph{} also achieves the highest mean
ARI and NMI for expression-derived domain recovery across evaluated
clustering resolutions, with higher NMI than BLEEP on 84\% of held-out
slides, although absolute agreement remains modest.

Our experiments address five research questions:
\textbf{RQ1 (Performance):} How accurately does \morph{} predict
gene expression within and across represented organs?
\textbf{RQ2 (Transfer):} How well does it generalize to an unseen organ?
\textbf{RQ3 (Ablation):} How local, slide, and spatial-neighbor context
affect performance?
\textbf{RQ4 (Coordinate invariance):} How stable are predictions under
coordinate-frame transformations with image features fixed?
\textbf{RQ5 (Structure):} How well do predictions recover
expression-derived spatial organization?

Our contributions are:
\begin{enumerate}[leftmargin=1.5em,itemsep=1pt,topsep=2pt]

    \item We introduce \morph{}, which augments a self-attention
    Transformer with two pathways applied in every layer: local
    spatial-$k$NN attention and attention-pooled slide context. We establish
    coordinate-frame invariance of the local pathway under stated assumptions.

    \item Through matched experiments across organs, gene panels, and image
    encoders, including held-out-organ transfer, we show that the added
    local and slide pathways improve prediction over the self-attention
    backbone, with paired tests showing that the local pathway benefits from
    spatially defined neighborhoods.

    \item We evaluate structure beyond average prediction accuracy through
    expression-derived spatial-domain recovery and selected cancer-marker
    expression patterns.
\end{enumerate}

\section{Method}

MIST maps frozen spot-level histology features and measured spot coordinates to a spatial gene
expression map. It has three stages: coordinate-invariant neighborhood construction, a layer that
augments a self-attention backbone with local and slide context, and supervised prediction
with a correlation-aware objective.

\subsection{Problem Setup}
For a slide with $N$ spots, let
$\mathbf{F}=[\mathbf{f}_1,\ldots,\mathbf{f}_N]^\top\in\mathbb{R}^{N\times d_{\mathrm{in}}}$
denote frozen pathology features and
$\mathbf{P}=[\mathbf{p}_1,\ldots,\mathbf{p}_N]^\top\in\mathbb{R}^{N\times2}$ the corresponding
spot coordinates. The target
$\mathbf{Y}\in\mathbb{R}^{N\times G}$ contains log-normalized expression for $G$ genes. We learn
$f_\theta:(\mathbf{F},\mathbf{P})\mapsto\hat{\mathbf{Y}}$ and process each slide independently;
the model therefore supports a variable number of spots and does not require expression measurements at
test time. In the feature-matched main comparisons, $\mathbf{f}_i$ concatenates frozen UNI
\cite{chen2024uni} and CONCH \cite{lu2024conch} embeddings ($d_{\mathrm{in}}=1536$).

\subsection{Coordinate-Invariant Neighborhood Construction}
We compute pairwise Euclidean distances $D_{ij}=\lVert\mathbf{p}_i-\mathbf{p}_j\rVert_2$ and use
them to select the indices $\mathcal{N}(i)$ of each spot's $k$ nearest neighbors, excluding the spot
itself. Coordinates enter the model only through this neighborhood graph: distances are not used as
attention features or biases, and absolute positions are not added to spot embeddings. For any
orthogonal transformation $R$ and translation $\mathbf{t}$, the transformed coordinates
$\mathbf{p}'_i=R\mathbf{p}_i+\mathbf{t}$ preserve every pairwise Euclidean distance. Therefore, with
consistent tie handling, they produce the same $k$NN graph and the same predictions for fixed spot
features. The formal guarantee is given in Proposition~\ref{prop:coordinate_invariance}. This is
\emph{coordinate-frame} invariance of the spatial pathway given fixed spot features; it is distinct
from invariance to physically rotating or flipping the H\&E image, which would change the frozen
encoder's embeddings $\mathbf{F}$ and is outside this guarantee.

\subsection{MIST Layer: Self-Attention Backbone with Local and Slide Context}

\insightbox{Core Idea.}{Each layer contains a self-attention backbone
over all spots and two added pathways: one that aggregates information from
each spot's spatial neighbors and one that summarizes the whole slide.
All three outputs are computed from the same normalized spot representations
and combined through a residual update. Only the local pathway uses
coordinates, and only to construct the spatial neighborhood.}

The input features are projected to hidden width $d$,
$\mathbf{X}^{(0)}=\mathbf{F}W_{\mathrm{in}}+\mathbf{b}_{\mathrm{in}}$.
The same spatial neighbor graph is reused in all $L$ Transformer layers.
At layer $\ell$, we first compute
\[
\mathbf{H}=\operatorname{LN}(\mathbf{X}^{(\ell)}),
\]
which is passed in parallel through the self-attention backbone and the
two added pathways.

\subsubsection{Self-attention (backbone)}
Standard multi-head self-attention \cite{vaswani2017attention} lets every spot attend to all $N$
spots. This backbone captures long-range dependencies but treats spots interchangeably: it introduces
no direct coordinate-dependent bias and is permutation-equivariant with respect to spot order. In
later layers, its inputs may nevertheless carry spatial information propagated by the two added
pathways.

\subsubsection{Local spatial-neighborhood attention (added)}
The first added pathway lets each spot attend to its $k$ neighbors using content-based attention. For
head $h$, the attention weights are
\begin{equation}
\alpha_{ij}^{h}=\operatorname{softmax}_{j\in\mathcal{N}(i)}\!\left(
\frac{(\mathbf{q}_i^{h})^\top\mathbf{k}_j^{h}}{\sqrt{d_h}}\right),
\end{equation}
where $d_h=d/H$ for $H$ heads. The local output is the concatenation of the weighted value sums,
followed by a learned output projection. Thus, coordinates determine the local neighborhood, while
the attention weights depend on the spot representations.

\subsubsection{Attention-pooled slide context (added)}
The second added pathway summarizes the slide. A learned scalar scoring function produces
$a_i=\operatorname{softmax}_{i}(s(\mathbf{h}_i))$. The slide representation is
\begin{equation}
\mathbf{g}=W_g\left(\sum_{i=1}^{N}a_i\mathbf{h}_i\right)+\mathbf{b}_g,
\end{equation}
and is broadcast to every spot. Unlike a learned class token, this representation is computed
directly from the slide's current spot representations and summarizes the slide-level context.

For the complete model, the backbone ($\mathbf{G}$) and the two added pathways, local
($\mathbf{L}$) and broadcast slide ($\mathbf{S}$), are summed and added to the residual stream. A
second pre-normalization residual update applies a two-layer feed-forward network with expansion
factor four and GELU activation:
\begin{align}
\widetilde{\mathbf{X}}^{(\ell)}
  &=\mathbf{X}^{(\ell)}+\mathbf{L}^{(\ell)}+\mathbf{G}^{(\ell)}+\mathbf{S}^{(\ell)},\\
\mathbf{X}^{(\ell+1)}
  &=\widetilde{\mathbf{X}}^{(\ell)}+
    \operatorname{FFN}\!\left(\operatorname{LN}(\widetilde{\mathbf{X}}^{(\ell)})\right).
\end{align}
After $L$ layers, layer normalization followed by a linear head maps each spot representation to its $G$ predicted gene values. Thus, \morph{} produces the complete slide-level gene expression prediction in a single forward pass.

\par\noindent\refstepcounter{proposition}\textbf{Proposition \theproposition\ (Coordinate frame invariance):}\label{prop:coordinate_invariance}\par
For fixed model parameters and fixed histology features, in deterministic evaluation mode, let
$R\in\mathbb{R}^{2\times2}$ be orthogonal, let $\mathbf{t}\in\mathbb{R}^{2}$, and define
$\mathbf{P}'=\mathbf{P}R^\top+\mathbf{1}\mathbf{t}^\top$. If the $k$NN construction resolves distance
ties consistently, the prediction is unchanged:
\begin{equation}
f_\theta(\mathbf{F},\mathbf{P}')=f_\theta(\mathbf{F},\mathbf{P}).
\end{equation}

\par\noindent\refstepcounter{proposition}\textbf{Proposition \theproposition\ (Permutation equivariance):}\label{prop:permutation_equivariance}\par
Let $Q\in\{0,1\}^{N\times N}$ be a permutation matrix. If the $k$NN construction handles distance
ties equivariantly, then
\begin{equation}
f_\theta(Q\mathbf{F},Q\mathbf{P})=Qf_\theta(\mathbf{F},\mathbf{P}).
\end{equation}
Therefore, reordering the input spots only reorders their predicted expression profiles.
Details, qualifications, and the proof for both of these propositions are provided in
Appendix~\ref{app:formal_guarantees}. Our numerical checks evaluate both
Propositions~\ref{prop:coordinate_invariance} and~\ref{prop:permutation_equivariance} under finite
precision.

\subsection{Training Objective and Ablations}
\label{sec:loss}
For a slide with measured expression $\mathbf{Y}$, we optimize mean squared error together with a
gene-wise Pearson-correlation term:
\begin{equation}
\mathcal{L}=\mathrm{MSE}(\hat{\mathbf{Y}},\mathbf{Y})
+ \lambda\left(1-\frac{1}{|\mathcal{V}|}\sum_{g\in\mathcal{V}}\rho_g\right),
\end{equation}
where $\rho_g$ is the Pearson correlation between predicted and measured expression across the $N$
spots, $\mathcal{V}$ is the set of genes with nonzero target variance on that slide, and
$\lambda=0.5$. MSE constrains the scale of the predictions, while the correlation term encourages
the model to reproduce spot-to-spot expression patterns.

The complete model uses the self-attention backbone together with the local and slide pathways. The
contribution of each added pathway is evaluated in the factorial ablation study reported in
Section~\ref{sec:design}, which compares the self-attention (G) backbone with the local (L) and slide-level
(S) pathways added on top.

\section{Experimental Setup}
\label{sec:setup}
\noindent\textbf{Dataset and Image Encoder Selection.}\quad
We evaluate six HEST-1k~\cite{jaume2024hest} cohorts:
IDC (breast), CCRCC (kidney), LUNG (lung), PAAD (pancreas),
PRAD (prostate), and SKCM (skin). We concatenate frozen
UNI~\cite{chen2024uni} and CONCH~\cite{lu2024conch} embeddings,
which achieve the highest average predictive performance among
the evaluated encoder configurations. Dataset and encoder details
are provided in Appendix~\ref{app:extended_setup}.

\noindent\textbf{Gene-panel selection.}\quad
We use a 50-gene HVG panel by default and compare it with 50-gene
DEG and HMHVG panels (Table~\ref{tab:panel_strategy}).
HVG selection ranks genes by expression variance; DEG selection
uses differential expression across Leiden clusters; and HMHVG
selection combines mean expression and variability after variance
filtering. We also evaluate 10- and 100-gene HVG panels, and a
200-gene HVG panel used only for the cancer-marker case study
(Section~\ref{sec:casestudy}), since several literature markers lie
outside the smaller panels.
All selection steps use training data only. For each LOOO fold,
the panel is recomputed using the five training organs, excluding
expression data from the held-out organ.

\noindent\textbf{Evaluation regimes.}\quad
We evaluate three settings:
\emph{intra-cohort}, with a separate model trained and tested within
each cohort;
\emph{pooled multi-organ}, with one model trained across organs
and tested on held-out slides from represented organs; and
\emph{leave-one-organ-out (LOOO)}, with training on five organs
and testing on the sixth.
Organ identity is not provided as an input.
All train--test splits are slide-disjoint. Because patient
identifiers are unavailable in our data pipeline, patient-level
disjointness cannot be verified; intra-cohort and pooled results
therefore measure slide-level generalization.

\noindent\textbf{Metrics and validation.}\quad
We report mean per-gene Pearson correlation coefficient (PCC)
as the primary metric, mean squared error (MSE) and mean absolute
error (MAE) for expression accuracy, and adjusted Rand index (ARI)
and normalized mutual information (NMI) for expression-derived
domain recovery. Checkpoints are selected using inner validation data and
evaluated once on the corresponding outer test fold. Results are
averaged across outer folds and three random seeds.

\noindent\textbf{Baselines and implementation.}\quad
We select one representative, publicly reproducible method per H\&E-to-ST modeling family: patch
regression (ST-Net~\cite{he2020stnet}), spatial graph modeling (Hist2ST~\cite{zeng2022hist2st}),
cross-modal retrieval (BLEEP~\cite{xie2023bleep}), and conditional generation
(STEM~\cite{zhu2025diffusion}). All are evaluated under matched features, panels,
splits, budgets, and model-selection rules. Unless otherwise stated, \morph{}
uses four layers, width 256, eight spatial neighbors, and 4.88M trainable
parameters. Full preprocessing, split, hyperparameter, and implementation
details are provided in Appendix~\ref{app:extended_setup}.

\section{Results}
\label{sec:results}

\subsection{RQ1: Predictive performance across diverse organs}
\label{sec:rq1}

We first ask whether \morph{} consistently predicts gene expression across diverse organs. For a fair comparison, all methods use frozen UNI+CONCH features with matched gene panels and data splits.

We begin with \emph{within-organ evaluation}, where each model is trained and tested within the same organ. Across the 50-gene HVG, DEG, and HMHVG panels, \morph{} achieves mean PCCs of $0.496$, $0.615$, and $0.650$, respectively, compared with BLEEP's $0.444$, $0.559$, and $0.587$ (Table~\ref{tab:panel_strategy}). Across HVG panel sizes, \morph{} reaches PCC@10/50/100 of $0.600/0.496/0.559$, leading all six cohorts at PCC@50 with an $11.6\%$ cohort-averaged gain over BLEEP; the 10- and 100-gene panel results are in Appendix~\ref{app:panelsize} (Table~\ref{tab:main}). It also achieves the lowest average MSE and MAE across all evaluated panel strategies and sizes.

The improvement is consistent across runs. Treating each cohort--seed run as the statistical unit ($n{=}18$), \morph{} improves PCC@50 over BLEEP by $0.051$ (95\% bootstrap CI $[0.041,0.062]$; Wilcoxon $p=7.6\times10^{-6}$). At the gene level, \morph{} also achieves higher PCC than BLEEP for $87\%$ of the 300 evaluated genes; we report this descriptively because genes within a slide are correlated. Statistical details are provided in Appendix~\ref{app:significance}.

Finally, in the \emph{pooled multi-organ setting}, a single \morph{} model is trained jointly across all six organs using a shared 14-gene leakage-safe panel. It achieves the highest mean PCC ($0.599$), compared with BLEEP ($0.587$) and Hist2ST ($0.556$) (Table~\ref{tab:pooled6}). Together, these results show consistent gains both within individual organs and across a shared multi-organ setting.

\begin{table*}[t]
\centering
\setcounter{table}{0}
\caption{\textbf{Performance across gene-selection strategies (50-gene panels).} Dataset PCC is mean $\pm$ SD over three seeds; final columns are six-cohort mean PCC, MSE, and MAE. Best per strategy in \best{bold}; $\uparrow$ higher and $\downarrow$ lower is better.}
\label{tab:panel_strategy}
\setlength{\tabcolsep}{3pt}
\renewcommand{\arraystretch}{1.0}
\resizebox{1.00\linewidth}{!}{%
\begin{tabular}{llrrrrrrrrr}
\toprule
Panel & Model & IDC & CCRCC & LUNG & PAAD & PRAD & SKCM & PCC Avg. $\uparrow$ & MSE Avg. $\downarrow$ & MAE Avg. $\downarrow$ \\
\midrule
\multirow{5}{*}{\textbf{HVG}} & ST-Net & $0.551{\pm}0.01$ & $0.163{\pm}0.02$ & $0.521{\pm}0.01$ & $0.366{\pm}0.03$ & $0.287{\pm}0.02$ & $0.661{\pm}0.01$ & $0.425$ & $1.995$ & $0.969$ \\
 & Hist2ST & $0.305{\pm}0.11$ & $0.180{\pm}0.03$ & $0.430{\pm}0.05$ & $0.248{\pm}0.04$ & $0.323{\pm}0.01$ & $0.523{\pm}0.00$ & $0.335$ & $1.561$ & $0.890$ \\
 & BLEEP & $0.519{\pm}0.00$ & $0.175{\pm}0.02$ & $0.565{\pm}0.01$ & $0.412{\pm}0.03$ & $0.346{\pm}0.01$ & $0.650{\pm}0.00$ & $0.444$ & $1.318$ & $0.819$ \\
 & STEM & $0.434{\pm}0.00$ & $0.118{\pm}0.01$ & $0.407{\pm}0.00$ & $0.233{\pm}0.02$ & $0.207{\pm}0.02$ & $0.461{\pm}0.02$ & $0.310$ & $1.747$ & $0.920$ \\
 & \morph{} (ours) & \best{$0.581{\pm}0.01$} & \best{$0.252{\pm}0.01$} & \best{$0.589{\pm}0.01$} & \best{$0.437{\pm}0.03$} & \best{$0.410{\pm}0.01$} & \best{$0.707{\pm}0.01$} & \best{$0.496$} & \best{$1.233$} & \best{$0.794$} \\
\midrule
\multirow{5}{*}{\textbf{DEG}} & ST-Net & $0.697{\pm}0.01$ & $0.312{\pm}0.01$ & $0.566{\pm}0.01$ & $0.425{\pm}0.03$ & $0.413{\pm}0.00$ & $0.752{\pm}0.01$ & $0.528$ & $3.336$ & $1.457$ \\
 & Hist2ST & $0.549{\pm}0.07$ & $0.335{\pm}0.01$ & $0.518{\pm}0.01$ & $0.352{\pm}0.01$ & $0.437{\pm}0.01$ & $0.656{\pm}0.00$ & $0.475$ & $1.757$ & $1.021$ \\
 & BLEEP & $0.731{\pm}0.01$ & $0.304{\pm}0.00$ & $0.621{\pm}0.01$ & $0.487{\pm}0.03$ & $0.466{\pm}0.00$ & $0.746{\pm}0.03$ & $0.559$ & $1.557$ & $0.945$ \\
 & STEM & $0.684{\pm}0.02$ & $0.253{\pm}0.01$ & $0.440{\pm}0.01$ & $0.306{\pm}0.04$ & $0.355{\pm}0.03$ & $0.617{\pm}0.02$ & $0.443$ & $1.958$ & $1.054$ \\
 & \morph{} (ours) & \best{$0.774{\pm}0.01$} & \best{$0.402{\pm}0.01$} & \best{$0.662{\pm}0.00$} & \best{$0.518{\pm}0.03$} & \best{$0.539{\pm}0.01$} & \best{$0.794{\pm}0.00$} & \best{$0.615$} & \best{$1.477$} & \best{$0.928$} \\
\midrule
\multirow{5}{*}{\textbf{HMHVG}} & ST-Net & $0.724{\pm}0.01$ & $0.339{\pm}0.01$ & $0.568{\pm}0.00$ & $0.441{\pm}0.03$ & $0.452{\pm}0.00$ & $0.824{\pm}0.03$ & $0.558$ & $3.589$ & $1.470$ \\
 & Hist2ST & $0.633{\pm}0.08$ & $0.387{\pm}0.01$ & $0.598{\pm}0.00$ & $0.341{\pm}0.04$ & $0.477{\pm}0.01$ & $0.710{\pm}0.14$ & $0.524$ & $1.832$ & $1.042$ \\
 & BLEEP & $0.739{\pm}0.02$ & $0.349{\pm}0.01$ & $0.661{\pm}0.00$ & $0.493{\pm}0.02$ & $0.432{\pm}0.04$ & $0.849{\pm}0.02$ & $0.587$ & $1.566$ & $0.936$ \\
 & STEM & $0.687{\pm}0.01$ & $0.290{\pm}0.01$ & $0.517{\pm}0.01$ & $0.305{\pm}0.03$ & $0.337{\pm}0.02$ & $0.731{\pm}0.06$ & $0.478$ & $1.976$ & $1.049$ \\
 & \morph{} (ours) & \best{$0.774{\pm}0.01$} & \best{$0.454{\pm}0.02$} & \best{$0.693{\pm}0.00$} & \best{$0.524{\pm}0.02$} & \best{$0.555{\pm}0.02$} & \best{$0.901{\pm}0.00$} & \best{$0.650$} & \best{$1.393$} & \best{$0.901$} \\
\bottomrule
\end{tabular}%
}
\end{table*}

\setcounter{table}{1}

\FloatBarrier

\begin{table}[!ht]
\centering
\caption{\textbf{Cross-organ POOLED (six-organ subset).} 5-fold slide-level holdout; mean$_{\text{std}}$ over three seeds; best in \best{bold}.}
\label{tab:pooled6}
{\small
\setlength{\tabcolsep}{6pt}
\begin{tabular}{l c}
\toprule
Method & PCC $\uparrow$ \\
\midrule
ST-Net~\cite{he2020stnet} (NBE'20)       & $0.351_{0.02}$ \\
Hist2ST~\cite{zeng2022hist2st} (BIB'22)     & $0.556_{0.01}$ \\
BLEEP~\cite{xie2023bleep} (NeurIPS'23)       & $0.587_{0.00}$ \\
STEM~\cite{zhu2025diffusion} (ICLR'25)             & $0.273_{0.00}$ \\
\midrule
\morph{} (ours)                          & \best{$0.599_{0.01}$} \\
\bottomrule
\end{tabular}
}
\end{table}

\subsection{RQ2: Cross-organ transfer to an unseen organ}
\label{sec:transfer}

Having established consistent performance across organs represented during training, we next ask whether \morph{} can transfer to an entirely unseen organ. We evaluate this using leave-one-organ-out (LOOO) evaluation, where each fold withholds one organ from training and tests exclusively on that organ, ensuring that its anatomy is absent from the training set. As shown in Table~\ref{tab:looo_6organ}, \morph{} achieves the best mean PCC ($0.422$ vs.\ BLEEP $0.398$) and matches or exceeds BLEEP on all six held-out organs, with PAAD effectively tied ($0.499$ vs.\ $0.497$). Across the 18 organ$\times$seed runs, this improvement is significant (two-sided Wilcoxon $p{=}0.018$; Appendix~\ref{app:significance}), showing that \morph{} retains consistent gains even when evaluated on organs excluded from training.

\begin{table}[!ht]
\centering
\caption{\textbf{Leave-one-organ-out (LOOO) performance.} Per-organ PCC, mean $\pm$ std over three seeds; final columns average PCC/MSE/MAE across the six held-out organs.}
\label{tab:looo_6organ}
\setlength{\tabcolsep}{3pt}
\resizebox{1.00\linewidth}{!}{%
\begin{tabular}{lrrrrrrrrr}
\toprule
Model & IDC & CCRCC & LUNG & PAAD & PRAD & SKCM & PCC Avg. $\uparrow$ & MSE Avg. $\downarrow$ & MAE Avg. $\downarrow$ \\
\midrule
ST-Net & $0.332{\pm}0.02$ & $0.074{\pm}0.00$ & $0.535{\pm}0.07$ & $0.382{\pm}0.00$ & $0.078{\pm}0.01$ & $0.500{\pm}0.09$ & $0.317$ & $1.912$ & $0.984$ \\
Hist2ST & $0.350{\pm}0.00$ & $0.064{\pm}0.01$ & $0.581{\pm}0.00$ & $0.214{\pm}0.17$ & $0.062{\pm}0.01$ & $0.694{\pm}0.02$ & $0.328$ & $1.356$ & $0.876$ \\
BLEEP & $0.401{\pm}0.04$ & $0.058{\pm}0.00$ & $0.630{\pm}0.00$ & $0.497{\pm}0.03$ & $0.091{\pm}0.00$ & $0.708{\pm}0.01$ & $0.398$ & $1.410$ & $0.847$ \\
STEM & $0.353{\pm}0.00$ & $0.055{\pm}0.02$ & $0.543{\pm}0.02$ & $0.393{\pm}0.02$ & $0.041{\pm}0.02$ & $0.492{\pm}0.07$ & $0.313$ & $1.413$ & \best{$0.814$} \\
\midrule
\morph{} (ours) & \best{$0.425{\pm}0.03$} & \best{$0.098{\pm}0.01$} & \best{$0.687{\pm}0.02$} & \best{$0.499{\pm}0.03$} & \best{$0.105{\pm}0.01$} & \best{$0.720{\pm}0.03$} & \best{$0.422$} & \best{$1.288$} & $0.849$ \\
\bottomrule
\end{tabular}%
}
\end{table}

\subsection{RQ3: How local, slide, and spatial-neighbor context affect performance}
\label{sec:design}

\begin{figure*}[t]
\centering
\includegraphics[width=1.00\linewidth]{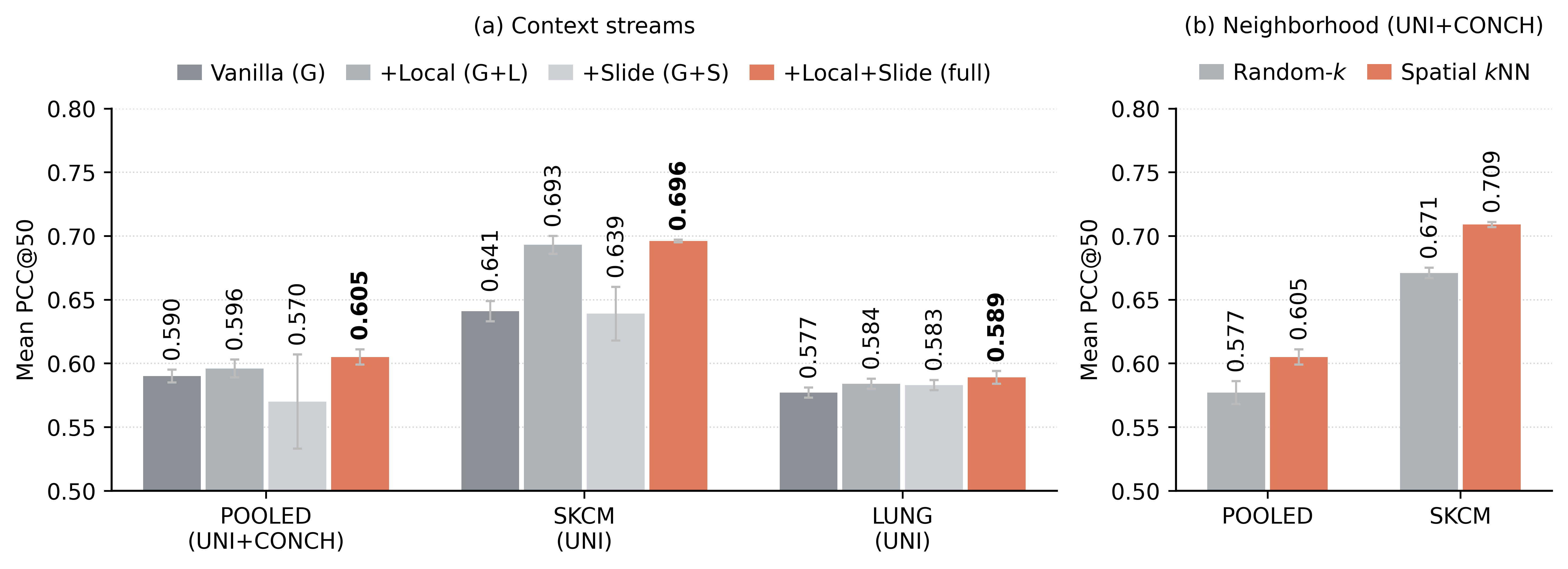}
\caption{\textbf{Context-stream ablation: self-attention is the vanilla Transformer backbone; \emph{local} and \emph{slide} are the added levers.} \textbf{(a)} Context Streams. \textbf{(b)} Neighborhood.}
\label{fig:ablation}
\end{figure*}

Since self-attention treats all spots interchangeably, we add two pathways on top
of this backbone, a coordinate-invariant local spatial-$k$NN pathway
(\mbox{+Local}) and a slide-pooling pathway (\mbox{+Slide}), and measure each
contribution on the pooled (UNI+CONCH) and SKCM/LUNG intra-cohort (UNI) settings,
pairing runs by (setting, seed, fold) and testing each gap with a two-sided
Wilcoxon signed-rank test (Figure~\ref{fig:ablation}). Local context alone
improves over the backbone ($+0.016$, $p{=}0.02$), and adding both pathways is
best in every setting ($0.605$ pooled, $0.696$ SKCM, $0.589$ LUNG; $+0.024$,
$p{=}3.4\times10^{-4}$, $n{=}27$); the two are complementary, as slide context
does not help alone ($-0.010$, $p{=}0.59$) but adds $+0.007$ on top of local
context. The local pathway relies on \emph{spatially} defined neighbors, not
merely extra attention: replacing $k$NN with random neighbors of the same size
costs $0.031$ PCC (Figure~\ref{fig:ablation}b; $p{=}8.4\times10^{-5}$, $n{=}21$).
Encoder-feature ablations are in Appendix~\ref{app:encoder}.

\subsection{RQ4: Does coordinate invariance hold in practice?}
\label{sec:coordinate_invariance}
\begin{table}[!ht]
\centering
\caption{\textbf{Coordinate-transform check on held-out slides.} PCC (mean $\pm$ std over three seeds, image features fixed); $\Delta$PCC relative to original coordinates. The pooled column uses a different gene panel.}
\label{tab:invariance}
\small
\setlength{\tabcolsep}{4pt}
\resizebox{1.00\linewidth}{!}{%
\begin{tabular}{l cc @{\hspace{12pt}} cc @{\hspace{12pt}} cc}
\toprule
 & \multicolumn{2}{c}{CCRCC (HVG; six folds)} & \multicolumn{2}{c}{PRAD (HVG; two folds)} & \multicolumn{2}{c}{Pooled (14 genes; five folds)} \\
\cmidrule(lr){2-3}\cmidrule(lr){4-5}\cmidrule(lr){6-7}
Coordinates & PCC $\uparrow$ & $\Delta$PCC & PCC $\uparrow$ & $\Delta$PCC & PCC $\uparrow$ & $\Delta$PCC \\
\midrule
Original         & $0.25158{\pm}0.01041$ & --- & $0.41020{\pm}0.00730$ & --- & $0.59875{\pm}0.00671$ & --- \\
Translation      & $0.25157{\pm}0.01027$ & $-0.000014$ & $0.41025{\pm}0.00732$ & $+0.000058$ & $0.59871{\pm}0.00668$ & $-0.000036$ \\
Rotation         & $0.25175{\pm}0.01053$ & $+0.000166$ & $0.41026{\pm}0.00729$ & $+0.000066$ & $0.59876{\pm}0.00670$ & $+0.000010$ \\
Reflection       & $0.25158{\pm}0.01041$ & $0$ & $0.41020{\pm}0.00730$ & $0$ & $0.59875{\pm}0.00671$ & $0$ \\
Spot permutation & $0.25154{\pm}0.01029$ & $-0.000035$ & $0.41022{\pm}0.00730$ & $+0.000026$ & $0.59870{\pm}0.00674$ & $-0.000049$ \\
\bottomrule
\end{tabular}}
\end{table}

In the earlier section, we showed that the added local and slide pathways improve prediction; we next test whether the model is stable to changes in the coordinate frame. With image features fixed, we translate spot coordinates by $(37.2,-91.4)$, rotate them by $0.731$ radians, reflect them across the vertical axis, and permute spot order. Across CCRCC, PRAD, and the pooled six-organ setting, these transformations cause only negligible changes in PCC(Table~\ref{tab:invariance}). Reflection leaves predictions unchanged, while the tiny differences under translation, rotation, and permutation arise from numerical effects in $k$NN selection when distances are tied or nearly tied.

\subsection{RQ5: Spatial structure preservation}
\label{sec:structure}

\begin{figure}[t]
\centering
\includegraphics[width=0.72\linewidth]{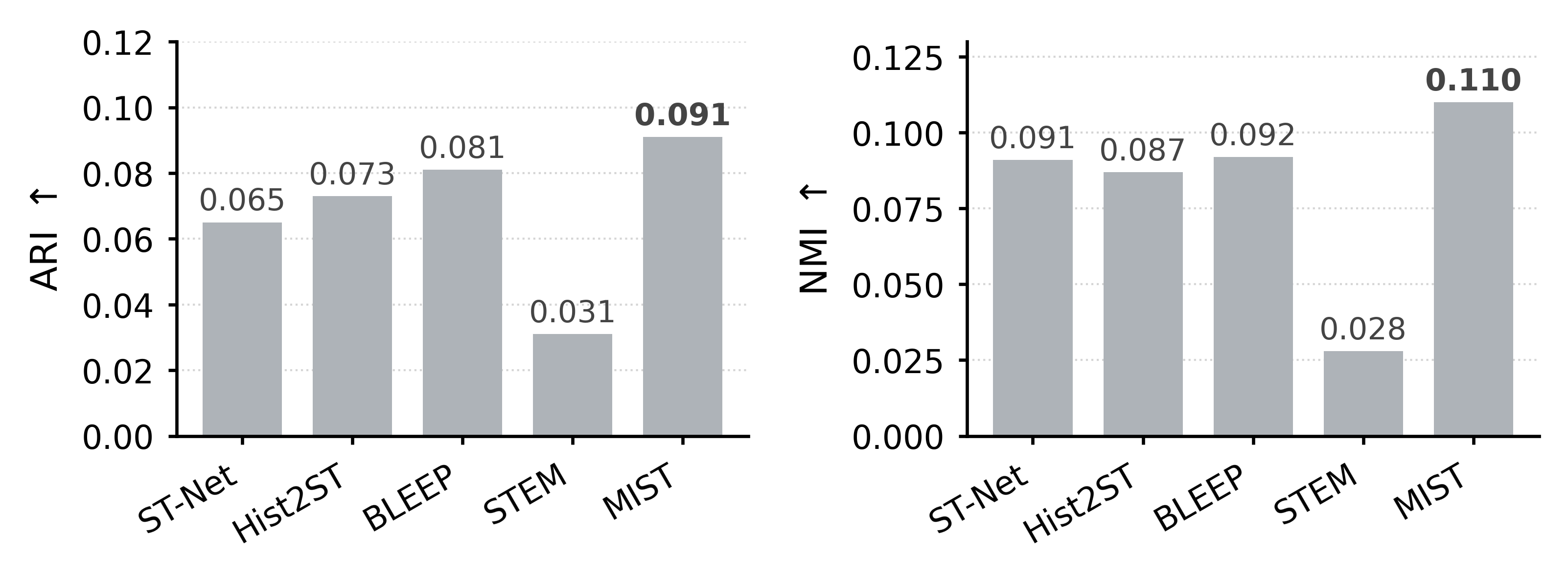}
\caption{\textbf{Spatial-domain recovery across all methods.} Mean ARI (left) and NMI (right)
(KMeans, $k{=}6$; predicted vs.\ measured domains); robustness to $k$ is in Appendix~\ref{app:krobust}.}
\label{fig:spatial_scatter}
\end{figure}

So far, we have shown that \morph{} predicts gene expression accurately. We next ask whether it also preserves spatial tissue structure. We cluster spots into $k{=}6$ domains using KMeans, separately from predicted and measured expression, and compare the resulting partitions using ARI and NMI on held-out slides (Figure~\ref{fig:spatial_scatter}).

\morph{} achieves the highest mean ARI ($0.091$) and NMI ($0.110$), outperforming BLEEP ($0.081$/$0.092$). This ordering remains unchanged for $k\in{4,6,8,10}$ (Appendix~\ref{app:krobust}). Although agreement is modest, \morph{} achieves higher NMI on $84\%$ of slides and higher ARI on $60\%$ of slides against BLEEP ($n{=}58$; $p{=}1.5\times10^{-8}$ and $0.03$, respectively). These results indicate that \morph{} better preserves spatial gene expression structure, particularly according to NMI.

\begin{figure*}[t]
\centering
\includegraphics[width=1.00\linewidth]{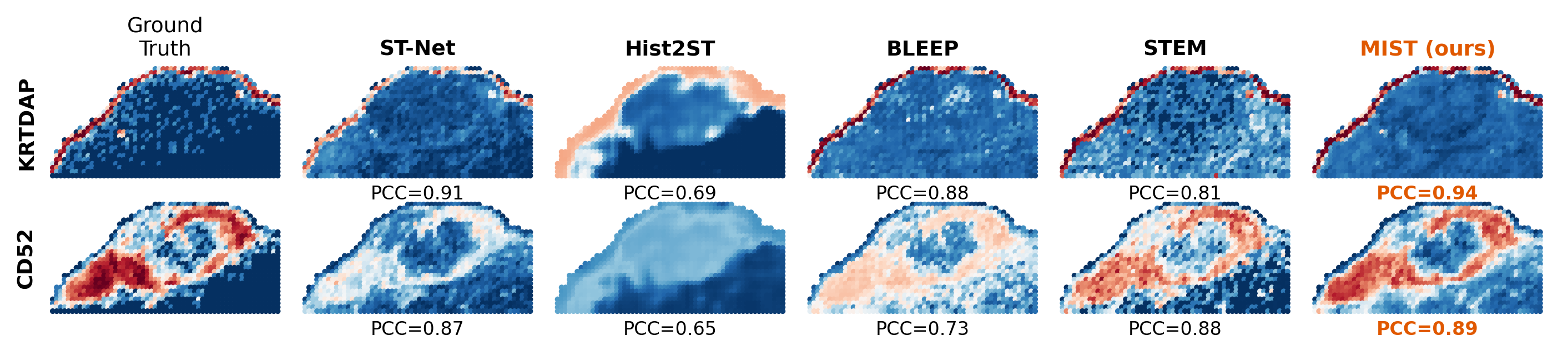}
\caption{\textbf{Qualitative spatial expression maps} for SKCM slide TENX117 (two marker genes,
rows); per-gene PCC shown, \morph{} in orange. }
\label{fig:qualitative}
\end{figure*}

\subsection{Case Study: Recovering Cancer-Marker Genes}
\label{sec:casestudy}

\begin{figure}[!ht]
\centering
\includegraphics[width=0.80\linewidth]{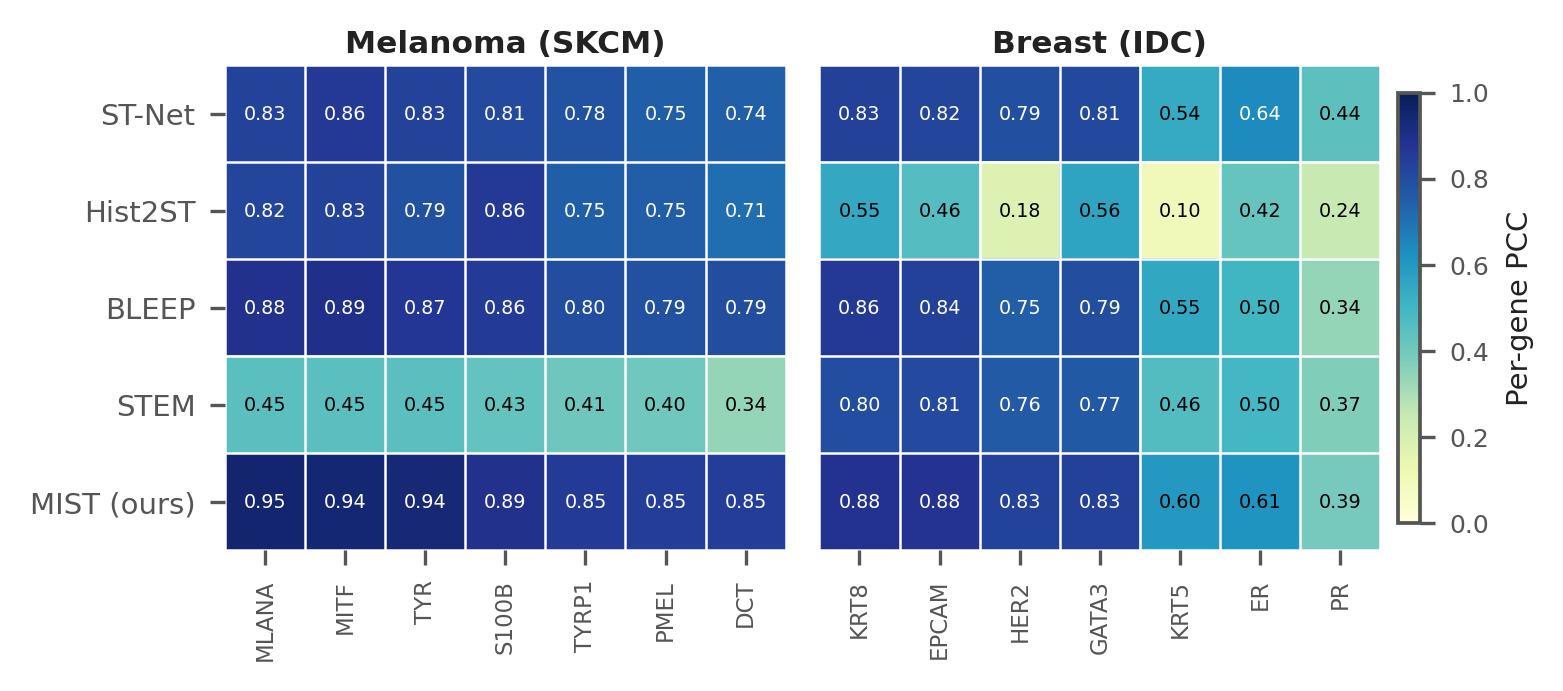}
\caption{\textbf{Case study: per-gene recovery of canonical cancer markers.} Per-gene Pearson
(200-gene panel, mean over folds and three seeds); darker is higher.}
\label{tab:casestudy}
\end{figure}

To further test its biological relevance, we ask whether \morph{} can recover canonical cancer-marker genes. We focus on established markers for melanoma (SKCM)~\cite{ohsie2008melanoma} and breast cancer (IDC), the latter including the clinically used HER2, ER, and PR receptor markers~\cite{wolff2018her2,allison2020erpr}. Because several markers are outside the default panel, we evaluate them using the 200-gene panel and measure the Pearson correlation between predicted and measured expression for each gene (Figure~\ref{tab:casestudy}).

\begin{figure}[!ht]
\centering
\includegraphics[width=1.00\linewidth]{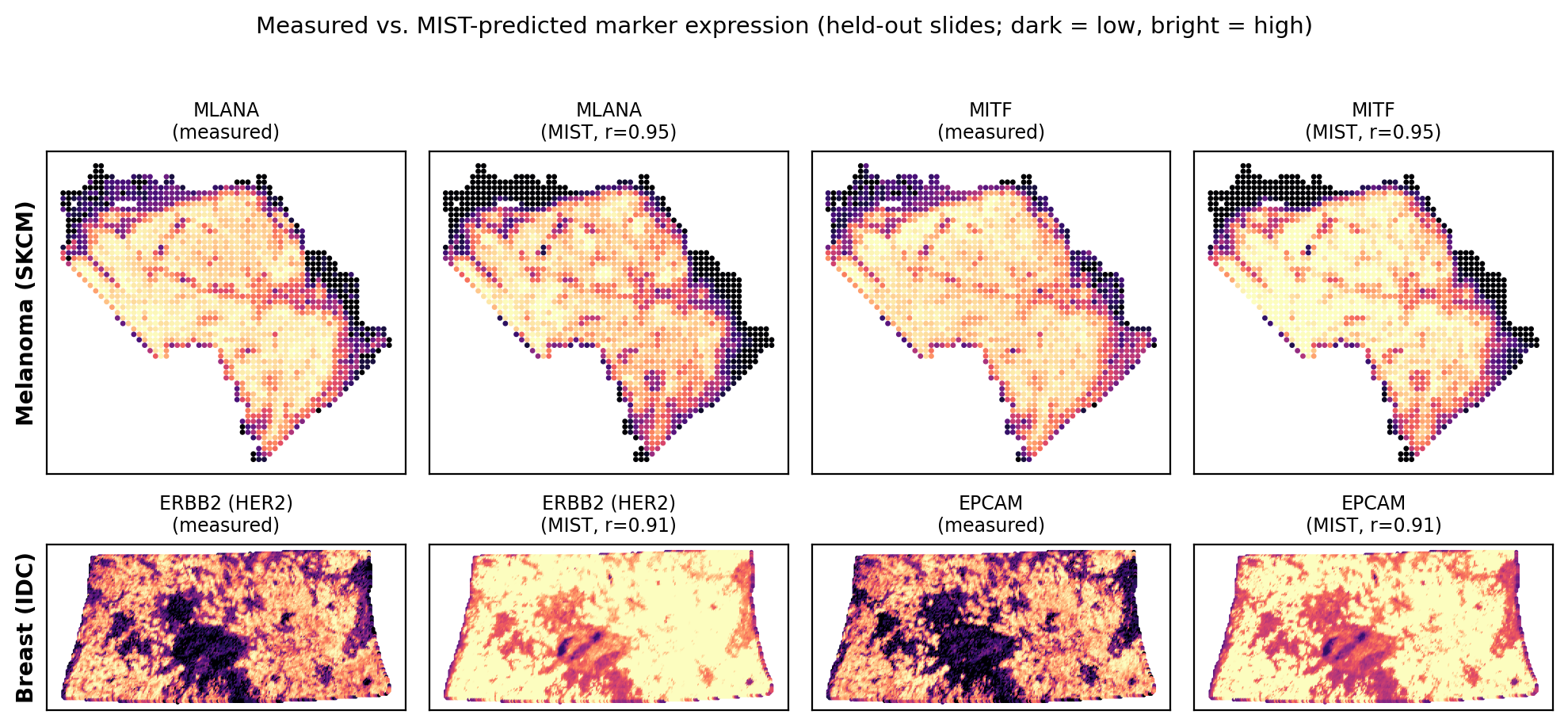}
\caption{\textbf{Spatial recovery of cancer markers.} Measured vs.\ \morph{}-predicted expression on
held-out melanoma (top) and breast (bottom) slides (dark = low, bright = high; per-panel $r$ shown).}
\label{fig:casestudy}
\end{figure}

\morph{} achieves high correlation for melanoma markers (mean $0.895$; MLANA $0.949$) and good correlation for breast-cancer markers (mean $0.717$; HER2 $0.829$). It obtains the highest average correlation for both cancers, with the top score for all melanoma markers and five of seven breast-cancer markers. Moreover, its predicted maps closely match the measured tumor layout (Figure~\ref{fig:casestudy}, $r\ge0.91$).

Together, these results show that \morph{} can recover both the expression and spatial patterns of several canonical cancer markers from H\&E. We interpret this as improved marker-gene prediction, not as evidence of standalone clinical utility.

\section{Related Work}
\noindent\textbf{H\&E-to-ST prediction.}\quad Existing methods include patch-level regression
(ST-Net~\cite{he2020stnet}), spatial graph modeling (Hist2ST~\cite{zeng2022hist2st}),
cross-modal retrieval (BLEEP~\cite{xie2023bleep}), and conditional diffusion
(STEM~\cite{zhu2025diffusion}). We evaluate these four baselines under matched
features, gene panels, splits, and training budgets.

\noindent\textbf{Context modeling.}\quad Spatial predictors differ in how they aggregate context:
ST-Net uses local image patches, while Hist2ST adds a graph over neighboring spots.
\morph{} combines content-based attention over spatial-$k$NN neighbors,
standard self-attention over all spots, and an attention-pooled tissue summary in every layer.

\noindent\textbf{Pathology foundation models.}\quad Recent pathology encoders provide transferable
representations for morphology and tissue composition~\cite{chen2024uni,lu2024conch}. We use frozen
UNI and CONCH features as matched inputs for \morph{} and the baselines, and separately evaluate
encoder choice to distinguish representation quality from architectural effects.

\noindent\textbf{Geometry and generalization.}\quad Absolute position encodings can change when an
equivalent tissue layout is translated, rotated, or reflected; geometric networks instead construct
interactions from relative quantities~\cite{satorras2021egnn}. \morph{} uses pairwise Euclidean
distances, which are invariant to these frame changes for fixed histology features.

\section{Conclusion and Future Work}

We introduced \morph{}, which augments a self-attention backbone with
local spatial-neighborhood attention and slide-level context for gene
expression prediction from histology. Across six HEST-1k organ cohorts,
\morph{} improves predictive performance, while ablations show that the gains
come from the added local and slide pathways and from using spatially defined
neighbors. The model also improves held-out-organ transfer and better preserves
expression-derived spatial structure. Generalization to unseen organs remains challenging. Future work will explore
slide-level metadata, transcriptome-wide prediction, end-to-end image encoding,
and predictive uncertainty to improve robustness and generalization.

\subsection*{Reproducibility statement}
All experiments use the public HEST-1k dataset~\cite{jaume2024hest}. Preprocessing, gene-panel
selection (including the leakage-safe per-fold panels for LOOO), train/test splits, evaluation
regimes, metrics, model hyperparameters, and baseline configurations are specified in
Section~\ref{sec:setup} and Appendix~\ref{app:extended_setup}; the complete model forward pass is
given in Algorithm~\ref{alg:mist}. Code to reproduce the model and the matched-baseline protocol is
available at an anonymized repository:
\url{https://anonymous.4open.science/r/MIST-F660/}.

\subsection*{Use of large language models}
Large language model (LLM) based tools were used during the preparation of this work only for:
(i)~language editing and proofreading of the manuscript; (ii)~assistance writing and refactoring
plotting, results-aggregation, and \LaTeX{} formatting scripts; and (iii)~consistency checks between
the reported tables and figures and the narrative. LLMs were not used to design the method, to
generate or modify experimental data, to run model training or evaluation, or to produce any reported
numerical result. All experiments were executed by the authors using the described pipeline, and the
authors reviewed all AI-assisted text take full responsibility for the content.

\bibliographystyle{iclr2027_conference}
\bibliography{references}

\clearpage
\appendix
\etocdepthtag.toc{supp}
\clearpage
\begin{center}
{\LARGE\scshape Appendix}\\[1.2em]
{\large\scshape Table of Contents}
\end{center}
\vspace{0.8em}
\etocsettagdepth{mainbody}{none}
\etocsettagdepth{supp}{subsection}
\etocsettocstyle{}{}
\begingroup
\hypersetup{linkcolor=black}
\tableofcontents
\endgroup
\clearpage

\section{Formal Guarantees}
\label{app:formal_guarantees}

This section gives the detailed statements and proofs of the coordinate-frame invariance and
permutation-equivariance properties used in the main text. The guarantees concern deterministic
evaluation with fixed model parameters and fixed histology features. In finite-precision computation,
negligible round-off differences may occur; nearest-neighbor distance ties must also be resolved
consistently.

\subsection{Coordinate-frame invariance}

\textbf{Proposition 1 (Coordinate frame invariance).}
For fixed model parameters and fixed histology features, in deterministic evaluation mode, let
$R\in\mathbb{R}^{2\times2}$ be orthogonal, let $\mathbf{t}\in\mathbb{R}^{2}$, and define
$\mathbf{P}'=\mathbf{P}R^\top+\mathbf{1}\mathbf{t}^\top$. If the $k$NN construction resolves distance
ties consistently, the prediction is unchanged:
\begin{equation}
f_\theta(\mathbf{F},\mathbf{P}')=f_\theta(\mathbf{F},\mathbf{P}).
\end{equation}

\textit{Proof.}
For any two spots,
\begin{align}
\mathbf{p}'_i-\mathbf{p}'_j
  &= (\mathbf{p}_i-\mathbf{p}_j)R^\top,\\
\|\mathbf{p}'_i-\mathbf{p}'_j\|_2^2
  &= (\mathbf{p}_i-\mathbf{p}_j)R^\top R(\mathbf{p}_i-\mathbf{p}_j)^\top
   = \|\mathbf{p}_i-\mathbf{p}_j\|_2^2.
\end{align}
Thus, the full distance matrix and the distance-based $k$NN ordering are unchanged, assuming
consistent tie handling. Any optional median-distance normalization is also unchanged because its
input distances are unchanged. The local attention receives the same neighbor indices, while the
self-attention, slide pooling, residual blocks, and prediction head receive the same histology
features and intermediate representations. Induction over the transformer layers gives identical
outputs. The result does not assert invariance of the frozen image encoder to transformations of the
H\&E image itself. $\square$

\subsection{Permutation equivariance}

\textbf{Proposition 2 (Permutation equivariance).}
Let $Q\in\{0,1\}^{N\times N}$ be a permutation matrix. If the $k$NN construction resolves distance
ties equivariantly, then
\begin{equation}
f_\theta(Q\mathbf{F},Q\mathbf{P})=Qf_\theta(\mathbf{F},\mathbf{P}).
\end{equation}

\textit{Proof.}
Permuting the rows of $\mathbf{P}$ permutes the rows and columns of its pairwise distance matrix.
With equivariant tie handling, the neighbor list and corresponding distances are permuted in the
same way. Hence each local attention computation is permuted with its associated spot. The
self-attention operation is permutation-equivariant because queries, keys, and values are permuted
together. Attention pooling produces the same pooled vector, since it is a weighted sum over the
spots, while broadcasting that vector preserves the row permutation. The input projection, residual
addition, layer normalization, feed-forward network, and final per-spot prediction head are all
row-wise equivariant. Applying these observations inductively through all layers yields
$f_\theta(Q\mathbf{F},Q\mathbf{P})=Qf_\theta(\mathbf{F},\mathbf{P})$. $\square$

\section{MIST Algorithm}
\label{app:algorithm}
Algorithm~\ref{alg:mist} gives the full forward pass corresponding to the Method:
coordinate-based $k$NN construction, $L$ multi-context layers that fuse local
$k$NN attention, self-attention, and an attention-pooled slide token, and a linear gene head.
Coordinates enter only through the neighbor graph, which is computed once and reused in every layer;
local attention weights are content-based (no distance features or positional biases).

\begin{algorithm}[H]
\caption{\morph{} forward pass (one slide)}
\label{alg:mist}
\begin{algorithmic}[1]
\Require frozen features $\mathbf{F}\in\mathbb{R}^{N\times d_{\mathrm{in}}}$, coordinates
  $\mathbf{P}\in\mathbb{R}^{N\times2}$; layers $L$, neighbors $k$
\Ensure predicted expression $\widehat{\mathbf{Y}}\in\mathbb{R}^{N\times G}$
\State $D_{ij}\gets \lVert \mathbf{p}_i-\mathbf{p}_j\rVert_2$ \Comment{pairwise distances (frame-invariant)}
\State $\mathcal{N}(i) \gets k\text{NN}(D)$ \Comment{neighbor indices, self excluded (only use of coordinates)}
\State $\mathbf{X}\gets \mathbf{F}\mathbf{W}_{\mathrm{in}}+\mathbf{b}_{\mathrm{in}}$ \Comment{input projection to width $d$}
\For{$\ell=1$ \textbf{to} $L$}
  \State $\mathbf{H}\gets \operatorname{LN}(\mathbf{X})$
  \State $\mathbf{L}\gets \textsc{LocalKNNAttn}(\mathbf{H},\mathcal{N})$ \Comment{content-based attention over $k$ neighbors}
  \State $\mathbf{G}\gets \textsc{SelfAttn}(\mathbf{H})$ \Comment{full self-attention over all $N$ spots}
  \State $a_i\gets \operatorname{softmax}_i(s(\mathbf{h}_i))$;\quad
         $\mathbf{g}\gets \mathbf{W}_g\!\sum_i a_i\mathbf{h}_i+\mathbf{b}_g$ \Comment{attention-pooled slide token}
  \State $\mathbf{X}\gets \mathbf{X}+\mathbf{L}+\mathbf{G}+\mathbf{1}\mathbf{g}^\top$ \Comment{parallel fusion + residual}
  \State $\mathbf{X}\gets \mathbf{X}+\operatorname{FFN}(\operatorname{LN}(\mathbf{X}))$
\EndFor
\State $\widehat{\mathbf{Y}}\gets \operatorname{LN}(\mathbf{X})\mathbf{W}_{\mathrm{head}}+\mathbf{b}_{\mathrm{head}}$
\State \Return $\widehat{\mathbf{Y}}$
\end{algorithmic}
\end{algorithm}

\noindent Training minimizes
$\mathcal{L}=\mathrm{MSE}(\widehat{\mathbf{Y}},\mathbf{Y})+\lambda\big(1-\tfrac{1}{|\mathcal{V}|}\sum_{g\in\mathcal{V}}\rho_g\big)$
with $\lambda{=}0.5$, where $\rho_g$ is the per-gene Pearson correlation over spots and $\mathcal{V}$
is the set of genes with nonzero target variance on the slide.

\section{Extended Experimental Setup}
\label{app:extended_setup}

\paragraph{Dataset and cohort selection.}
We use \textbf{HEST-1k}~\cite{jaume2024hest}, a collection of spatial transcriptomics
measurements paired with matched H\&E whole-slide images, available from the
\href{https://huggingface.co/datasets/MahmoodLab/hest}{HEST-1k Hugging Face repository}. HEST-1k has
supported subsequent histology-to-transcriptomics studies, including STEM and STPath
\cite{zhu2025diffusion,stpath2025}. Our experiments use six cohorts: CCRCC, IDC, LUNG, PAAD, PRAD,
and SKCM. Appendix~\ref{app:cohort_stats} reports their organs, sample and spot counts, and fold
counts.

\paragraph{Image encoders.}
We use frozen UNI (ViT-L/16, 1,024-dimensional image embeddings;
\href{https://huggingface.co/MahmoodLab/UNI}{Hugging Face model card})~\cite{chen2024uni} and
CONCH (ViT-B/16, 512-dimensional image embeddings;
\href{https://huggingface.co/MahmoodLab/CONCH}{Hugging Face model card})~\cite{lu2024conch}.
Both encoders were evaluated for gene expression prediction in the HEST-1k benchmark
\cite{jaume2024hest}. We test UNI and CONCH separately and concatenate their features to obtain
1,536-dimensional spot embeddings. Concatenation gives the highest mean PCC in our encoder
comparison: 0.496, compared with 0.484 for UNI and 0.463 for CONCH alone (Table~\ref{tab:main};
Appendix~\ref{tab:encoder}). We therefore use UNI+CONCH in the main comparisons.

\paragraph{Preprocessing and gene panels.}
We use the provided slide-level cross-validation folds and transform expression targets with $\log(1+x)$.
Unless stated otherwise, targets are the 50-gene HEST HVG panel. The panel study additionally uses
matched 50-gene DEG, selected by Wilcoxon testing across Leiden clusters, and HMHVG panels, selected
using mean-expression and variability criteria after variance filtering (Table~\ref{tab:panel_strategy}).

\paragraph{Evaluation protocol.}
Intra-cohort training fits a model within each cohort. Pooled five-fold validation trains across
organs while holding out slide-level samples; organ identity defines the protocol but is not a model
input. LOOO trains on all-but-one organ and tests the held-out organ. We report PCC, MSE, MAE, and
(for domain recovery) ARI and NMI, averaged over outer folds and three seeds. Checkpoints are selected on inner validation and evaluated
once on the outer test fold; single-slide folds use disjoint spots for inner validation. Patient-level
disjointness is not guaranteed where identifiers are unavailable.

\paragraph{Baselines and implementation.}
We compare with ST-Net \cite{he2020stnet}, Hist2ST \cite{zeng2022hist2st}, BLEEP \cite{xie2023bleep},
and STEM \cite{zhu2025diffusion} under matched features, panels, splits, budgets, and checkpoint
selection. All methods use the \emph{same} fold assignments, the same frozen UNI+CONCH inputs, and the
same three random seeds ($1,2,3$); every method is trained per cohort/regime and its checkpoint is
selected by best inner-validation PCC, then evaluated once on the outer test fold. \morph{} uses four
layers, width 256, four heads, eight neighbors, and 4.88M parameters.
AdamW training uses learning rate $10^{-3}$, weight decay $10^{-2}$, cosine decay, gradient clipping,
100 epochs maximum, patience 20, and at most 3,000 spots per step.

\paragraph{System and software environment.}
All experiments run on Linux (RHEL~9, kernel~5.14) with Python~3.13 and PyTorch~2.12 (CUDA~13.0,
cuDNN~9.2). Core scientific dependencies are NumPy~2.4, SciPy~1.17, scikit-learn~1.8, pandas~2.3, and
scanpy~1.12. Frozen UNI and CONCH features are extracted with \texttt{timm} and the official model
weights. Training and inference use NVIDIA GPUs with compute capability $\geq$~sm\_80 (A100~80GB,
A100-PCIe, L40s, and H200; Appendix~\ref{app:compute}); Volta (V100) GPUs are incompatible with this
CUDA-13 build and were excluded. Cluster nodes use Intel Xeon Gold~6140 CPUs.

\section{Dataset}
\label{app:cohort_stats}

To evaluate \morph{}, we use six cohorts from the HEST-1k dataset~\cite{jaume2024hest}, a curated
collection of 10x Genomics Visium spatial transcriptomics samples paired with matched H\&E whole-slide
images (WSIs) and standardized metadata. Each cohort covers a distinct organ:
(1)~\textbf{CCRCC} (kidney) provides 24 WSIs and 74{,}220 spots;
(2)~\textbf{IDC} (breast) provides 4 WSIs and 44{,}974 spots;
(3)~\textbf{PRAD} (prostate) provides 23 WSIs and 62{,}710 spots;
(4)~\textbf{PAAD} (pancreas) provides 3 WSIs and 9{,}793 spots;
(5)~\textbf{LUNG} provides 2 WSIs and 7{,}505 spots; and
(6)~\textbf{SKCM} (skin) provides 2 WSIs and 5{,}716 spots.
Spots are counted after barcode filtering; all cohorts use Visium spots with a 55\,\textmu m diameter.
For each spot we use the frozen spot-level UNI+CONCH image embedding provided at the HEST-bench
resolution, and the $\log(1+x)$-transformed expression over the selected gene panel as the prediction
target. Per-cohort sample, spot, and cross-validation-fold counts are summarized in
Table~\ref{tab:cohort_stats}.

\begin{table}[H]
\centering
\caption{\textbf{HEST-1k cohort statistics.} Cohorts used in the reported experiments; spots are counted after barcode filtering and folds denote per-cohort train/test splits.}
\label{tab:cohort_stats}
\small
\begin{tabular}{l l r r r}
\toprule
Cohort & Organ & Samples & Spots & Folds \\
\midrule
IDC          & Breast      & 4 & 44{,}974 & 4 \\
CCRCC        & Kidney      & 24 & 74{,}220 & 6 \\
LUNG         & Lung        &  2 &  7{,}505 & 2 \\
PAAD         & Pancreas    &  3 &  9{,}793 & 3 \\
PRAD         & Prostate    & 23 & 62{,}710 & 2 \\
SKCM         & Skin        &  2 &  5{,}716 & 2 \\
\bottomrule
\end{tabular}
\end{table}

\paragraph{Split granularity and patient disjointness.}
Each ``sample'' in Table~\ref{tab:cohort_stats} is a distinct HEST-1k slide (a whole-slide section),
and every fold is \emph{slide-disjoint}: no slide is shared between training and test, the pooled
regime holds out entire slides, and the leave-one-organ-out regime holds out entire organs. HEST-1k
patient/donor identifiers are not distributed with the per-slide data in our pipeline, so for a cohort
in which one donor may contribute several slides we cannot verify patient-level disjointness. The
intra-cohort and pooled numbers should therefore be read as \emph{slide-level} generalization and
could be optimistic if a cohort mixes multiple slides from the same patient; this risk is largest for
the two multi-slide cohorts, CCRCC (24 slides) and PRAD (23 slides), and small for the remaining
cohorts (IDC~4, PAAD~3, LUNG~2, SKCM~2 slides). Importantly, the cross-organ LOOO transfer results
(our central generalization claim) are unaffected, because an entire organ (and hence a disjoint set
of patients) is withheld at test time.

\section{Paired Statistical Tests}
\label{app:significance}
This section collects the paired significance results reported in the main text. Tests compare
\morph{} with the strongest relevant baseline under the same splits and evaluation protocol. The
table reports the available aggregate statistics; complete per-run or per-slide effect-size vectors
are not included in the manuscript.

\begin{table}[H]
\centering
\caption{\textbf{Summary of paired statistical tests.} All tests are two-sided Wilcoxon signed-rank
tests unless stated otherwise; $p$-values are reported as available from the matched evaluation.}
\label{tab:paired_tests}
\small
\setlength{\tabcolsep}{4pt}
\begin{tabular}{p{0.23\linewidth}p{0.27\linewidth}p{0.15\linewidth}p{0.25\linewidth}}
\toprule
Outcome & Comparison & Sample size & Reported result \\
\midrule
PCC@50 & \morph{} vs. BLEEP & $n=18$ paired runs & gain $0.051$; 95\% bootstrap CI $[0.041,0.062]$; $p=7.6\times10^{-6}$ \\
Per-gene PCC@50 & \morph{} vs. BLEEP & $n=300$ genes (descriptive) & \morph{} higher on 87\% \\
LOOO PCC & \morph{} vs. BLEEP & $n=18$ organ$\times$seed; also $n=6$ cohorts & $p=0.018$ ($n{=}18$); exact $p=0.031$, all six positive ($n{=}6$) \\
Spatial ARI & \morph{} vs. BLEEP & $n=58$ slides & \morph{} wins 60\%; $p=0.03$ \\
Spatial NMI & \morph{} vs. BLEEP & $n=58$ slides & \morph{} wins 84\%; $p=1.5\times10^{-8}$ \\
\bottomrule
\end{tabular}
\end{table}

\section{Panel-Size Results (10 and 100 genes)}
\label{app:panelsize}
Table~\ref{tab:main} reports intra-cohort performance on the 10- and 100-gene HVG panels,
complementing the 50-gene results in Table~\ref{tab:panel_strategy}. \morph{} attains the highest
six-cohort mean PCC at both sizes (PCC@10 $0.600$, PCC@100 $0.559$) with the lowest average MSE and MAE.

\begin{table}[H]
\centering
\caption{\textbf{Performance across HVG panel sizes (10 and 100 genes).} Dataset PCC is mean $\pm$ SD over three seeds; final columns are six-cohort mean PCC, MSE, and MAE. The 50-gene HVG results are in Table~\ref{tab:panel_strategy}.}
\label{tab:main}
\label{tab:main_g10}
\label{tab:main_g100}
\setlength{\tabcolsep}{3pt}
\renewcommand{\arraystretch}{1.0}
\resizebox{1.00\linewidth}{!}{%
\begin{tabular}{llrrrrrrrrr}
\toprule
Panel size & Model & IDC & CCRCC & LUNG & PAAD & PRAD & SKCM & PCC Avg. $\uparrow$ & MSE Avg. $\downarrow$ & MAE Avg. $\downarrow$ \\
\midrule
\multirow{5}{*}{\textbf{PCC@10}} & ST-Net & $0.666{\pm}0.01$ & $0.231{\pm}0.01$ & $0.702{\pm}0.01$ & $0.511{\pm}0.04$ & $0.442{\pm}0.03$ & $0.728{\pm}0.00$ & $0.547$ & $6.087$ & $1.908$ \\
 & Hist2ST & $0.636{\pm}0.01$ & $0.273{\pm}0.02$ & $0.519{\pm}0.02$ & $0.377{\pm}0.07$ & $0.459{\pm}0.02$ & $0.471{\pm}0.04$ & $0.456$ & $3.297$ & $1.374$ \\
 & BLEEP & $0.635{\pm}0.02$ & $0.231{\pm}0.02$ & $0.735{\pm}0.00$ & \best{$0.536{\pm}0.03$} & $0.434{\pm}0.03$ & $0.697{\pm}0.01$ & $0.545$ & $2.859$ & $1.243$ \\
 & STEM & $0.591{\pm}0.02$ & $0.170{\pm}0.00$ & $0.616{\pm}0.01$ & $0.393{\pm}0.04$ & $0.325{\pm}0.03$ & $0.598{\pm}0.01$ & $0.449$ & $3.181$ & $1.310$ \\
 & \morph{} (ours) & \best{$0.708{\pm}0.00$} & \best{$0.322{\pm}0.01$} & \best{$0.759{\pm}0.00$} & $0.519{\pm}0.05$ & \best{$0.524{\pm}0.02$} & \best{$0.769{\pm}0.01$} & \best{$0.600$} & \best{$2.620$} & \best{$1.211$} \\
\midrule
\multirow{5}{*}{\textbf{PCC@100}} & ST-Net & $0.611{\pm}0.01$ & $0.221{\pm}0.01$ & $0.595{\pm}0.00$ & $0.355{\pm}0.04$ & $0.327{\pm}0.01$ & $0.696{\pm}0.01$ & $0.468$ & $3.320$ & $1.354$ \\
 & Hist2ST & $0.473{\pm}0.07$ & $0.274{\pm}0.03$ & $0.434{\pm}0.03$ & $0.222{\pm}0.04$ & $0.341{\pm}0.03$ & $0.446{\pm}0.12$ & $0.365$ & $1.843$ & $1.011$ \\
 & BLEEP & $0.632{\pm}0.01$ & $0.247{\pm}0.02$ & $0.659{\pm}0.01$ & $0.399{\pm}0.04$ & $0.377{\pm}0.02$ & $0.696{\pm}0.01$ & $0.502$ & $1.558$ & $0.915$ \\
 & STEM & $0.565{\pm}0.01$ & $0.192{\pm}0.00$$^\dag$ & $0.498{\pm}0.03$ & $0.236{\pm}0.02$ & $0.240{\pm}0.03$ & $0.602{\pm}0.02$ & $0.389$ & $1.899$ & $1.001$ \\
 & \morph{} (ours) & \best{$0.682{\pm}0.01$} & \best{$0.328{\pm}0.02$} & \best{$0.688{\pm}0.00$} & \best{$0.435{\pm}0.03$} & \best{$0.450{\pm}0.00$} & \best{$0.771{\pm}0.01$} & \best{$0.559$} & \best{$1.458$} & \best{$0.894$} \\
\bottomrule
\end{tabular}%
}
\end{table}

\section{Encoder Ablation}
\label{app:encoder}
We compare three encoder configurations on the intra-cohort PCC@50 task, holding models, cohorts, gene panels, splits, and training budget fixed: UNI only (ViT-L/16, 1024-d), CONCH only (512-d), and the concatenated UNI+CONCH representation (1536-d) used in the main results (Table~\ref{tab:encoder}). Concatenation is best for \morph{} (encoder-averaged PCC $0.463\!\to\!0.484\!\to\!0.496$ for CONCH, UNI, UNI+CONCH), and \morph{} remains the most accurate method under all three encoders.

\begin{table}[H]
\centering
\caption{\textbf{Encoder ablation: intra-cohort PCC@50 per encoder.} UNI-only (1024-d), CONCH-only (512-d), and concatenated UNI+CONCH (1536-d); mean$_{\text{std}}$ over three seeds; best per row in \best{bold}.}
\label{tab:encoder}
\setlength{\tabcolsep}{3pt}\renewcommand{\arraystretch}{1.0}
\resizebox{0.82\linewidth}{!}{%
\begin{tabular}{ll ccccc}
\toprule
Encoder & Cohort & ST-Net & Hist2ST & BLEEP & STEM & \morph{} (ours) \\
\midrule
\multirow{7}{*}{\rotatebox[origin=c]{90}{\textbf{\shortstack{UNI\\(1024-d)}}}} & IDC~\tiny{Breast} & $0.538{\pm}0.008$ & $0.250{\pm}0.067$ & $0.508{\pm}0.005$ & $0.436{\pm}0.005$ & \best{$0.569{\pm}0.009$} \\
 & CCRCC~\tiny{Kidney} & $0.172{\pm}0.010$ & $0.214{\pm}0.015$ & $0.184{\pm}0.007$ & $0.116{\pm}0.002$ & \best{$0.264{\pm}0.010$} \\
 & LUNG~\tiny{Lung} & $0.482{\pm}0.012$ & $0.378{\pm}0.097$ & $0.559{\pm}0.007$ & $0.393{\pm}0.009$ & \best{$0.590{\pm}0.005$} \\
 & PAAD~\tiny{Pancreas} & $0.353{\pm}0.022$ & $0.243{\pm}0.061$ & \best{$0.393{\pm}0.035$} & $0.227{\pm}0.017$ & $0.392{\pm}0.040$ \\
 & PRAD~\tiny{Prostate} & $0.251{\pm}0.008$ & $0.322{\pm}0.020$ & $0.343{\pm}0.009$ & $0.190{\pm}0.010$ & \best{$0.394{\pm}0.003$} \\
 & SKCM~\tiny{Skin} & $0.632{\pm}0.010$ & $0.510{\pm}0.015$ & $0.639{\pm}0.015$ & $0.471{\pm}0.031$ & \best{$0.694{\pm}0.001$} \\
\cmidrule(lr){2-7}
 & \textbf{Average} & $0.405$ & $0.319$ & $0.438$ & $0.305$ & \best{$0.484$} \\
\midrule
\multirow{7}{*}{\rotatebox[origin=c]{90}{\textbf{\shortstack{CONCH\\(512-d)}}}} & IDC~\tiny{Breast} & $0.515{\pm}0.007$ & $0.264{\pm}0.018$ & $0.496{\pm}0.011$ & $0.396{\pm}0.011$ & \best{$0.551{\pm}0.004$} \\
 & CCRCC~\tiny{Kidney} & $0.160{\pm}0.010$ & $0.126{\pm}0.026$ & $0.141{\pm}0.012$ & $0.078{\pm}0.001$ & \best{$0.222{\pm}0.010$} \\
 & LUNG~\tiny{Lung} & $0.517{\pm}0.001$ & $0.435{\pm}0.006$ & \best{$0.576{\pm}0.003$} & $0.386{\pm}0.030$ & $0.554{\pm}0.005$ \\
 & PAAD~\tiny{Pancreas} & $0.372{\pm}0.035$ & $0.186{\pm}0.035$ & $0.384{\pm}0.033$ & $0.251{\pm}0.028$ & \best{$0.387{\pm}0.045$} \\
 & PRAD~\tiny{Prostate} & $0.327{\pm}0.003$ & $0.322{\pm}0.009$ & $0.364{\pm}0.010$ & $0.210{\pm}0.013$ & \best{$0.406{\pm}0.006$} \\
 & SKCM~\tiny{Skin} & $0.633{\pm}0.002$ & $0.388{\pm}0.022$ & $0.633{\pm}0.005$ & $0.453{\pm}0.014$ & \best{$0.658{\pm}0.007$} \\
\cmidrule(lr){2-7}
 & \textbf{Average} & $0.421$ & $0.287$ & $0.432$ & $0.296$ & \best{$0.463$} \\
\midrule
\multirow{7}{*}{\rotatebox[origin=c]{90}{\textbf{\shortstack{UNI+CONCH\\(1536-d)}}}} & IDC~\tiny{Breast} & $0.551{\pm}0.01$ & $0.309{\pm}0.07$ & $0.519{\pm}0.00$ & $0.439{\pm}0.01$ & \best{$0.581{\pm}0.01$} \\
 & CCRCC~\tiny{Kidney} & $0.163{\pm}0.02$ & $0.184{\pm}0.02$ & $0.175{\pm}0.02$ & $0.123{\pm}0.00$ & \best{$0.252{\pm}0.01$} \\
 & LUNG~\tiny{Lung} & $0.518{\pm}0.01$ & $0.470{\pm}0.01$ & $0.562{\pm}0.01$ & $0.423{\pm}0.03$ & \best{$0.589{\pm}0.01$} \\
 & PAAD~\tiny{Pancreas} & $0.366{\pm}0.03$ & $0.248{\pm}0.04$ & $0.412{\pm}0.03$ & $0.233{\pm}0.02$ & \best{$0.437{\pm}0.03$} \\
 & PRAD~\tiny{Prostate} & $0.294{\pm}0.02$ & $0.333{\pm}0.02$ & $0.348{\pm}0.01$ & $0.207{\pm}0.02$ & \best{$0.410{\pm}0.01$} \\
 & SKCM~\tiny{Skin} & $0.665{\pm}0.01$ & $0.522{\pm}0.03$ & $0.645{\pm}0.01$ & $0.491{\pm}0.04$ & \best{$0.707{\pm}0.01$} \\
\cmidrule(lr){2-7}
 & \textbf{Average} & $0.425$ & $0.335$ & $0.444$ & $0.310$ & \best{$0.496$} \\
\bottomrule
\end{tabular}%
}
\end{table}

\begin{figure}[H]
\centering
\includegraphics[width=0.85\linewidth]{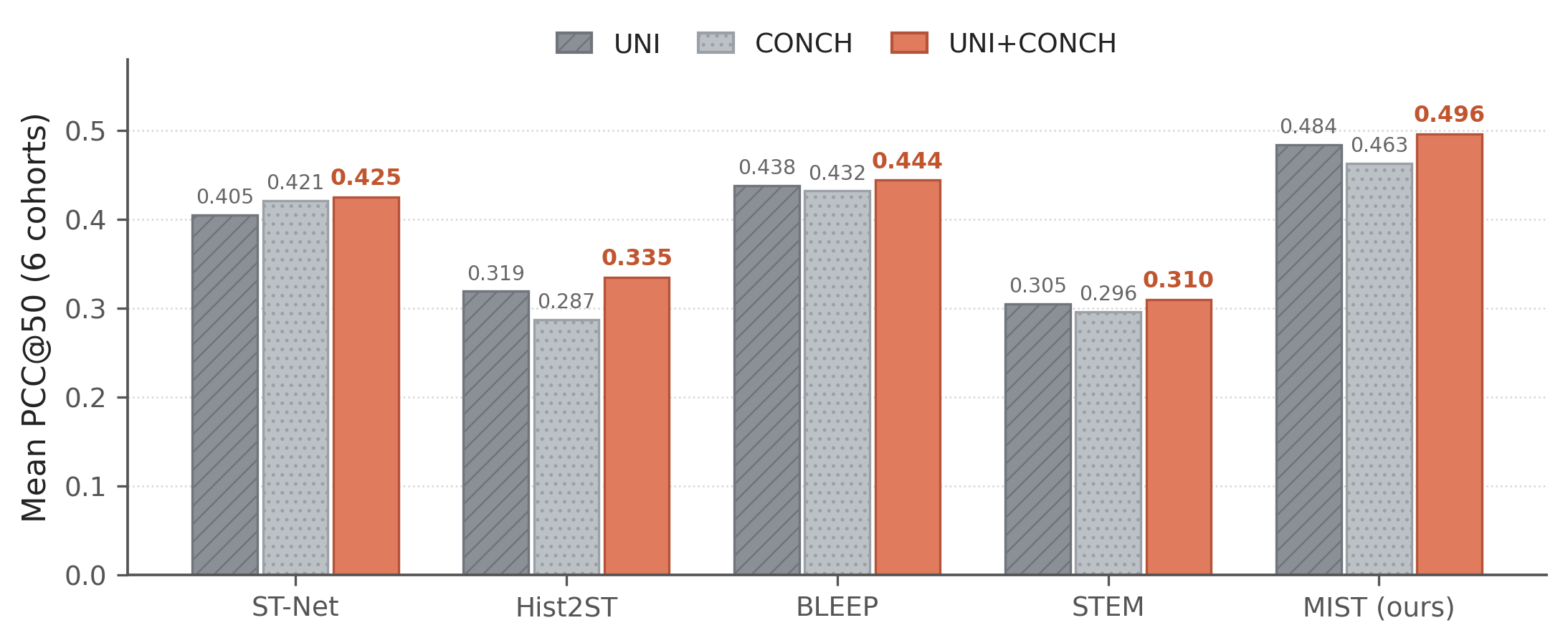}
\caption{\textbf{Encoder-averaged intra-cohort PCC@50 per model.} Six-cohort mean for each encoder
(UNI, CONCH, UNI+CONCH); same values as Table~\ref{tab:encoder}. \morph{} is highest under every
encoder, and UNI+CONCH is best for it.}
\label{fig:encoder_bars}
\end{figure}

\section{Cancer-Marker Recovery: All Cohorts}
\label{app:markers_supp}
\begin{figure}[H]
\centering
\includegraphics[width=\linewidth]{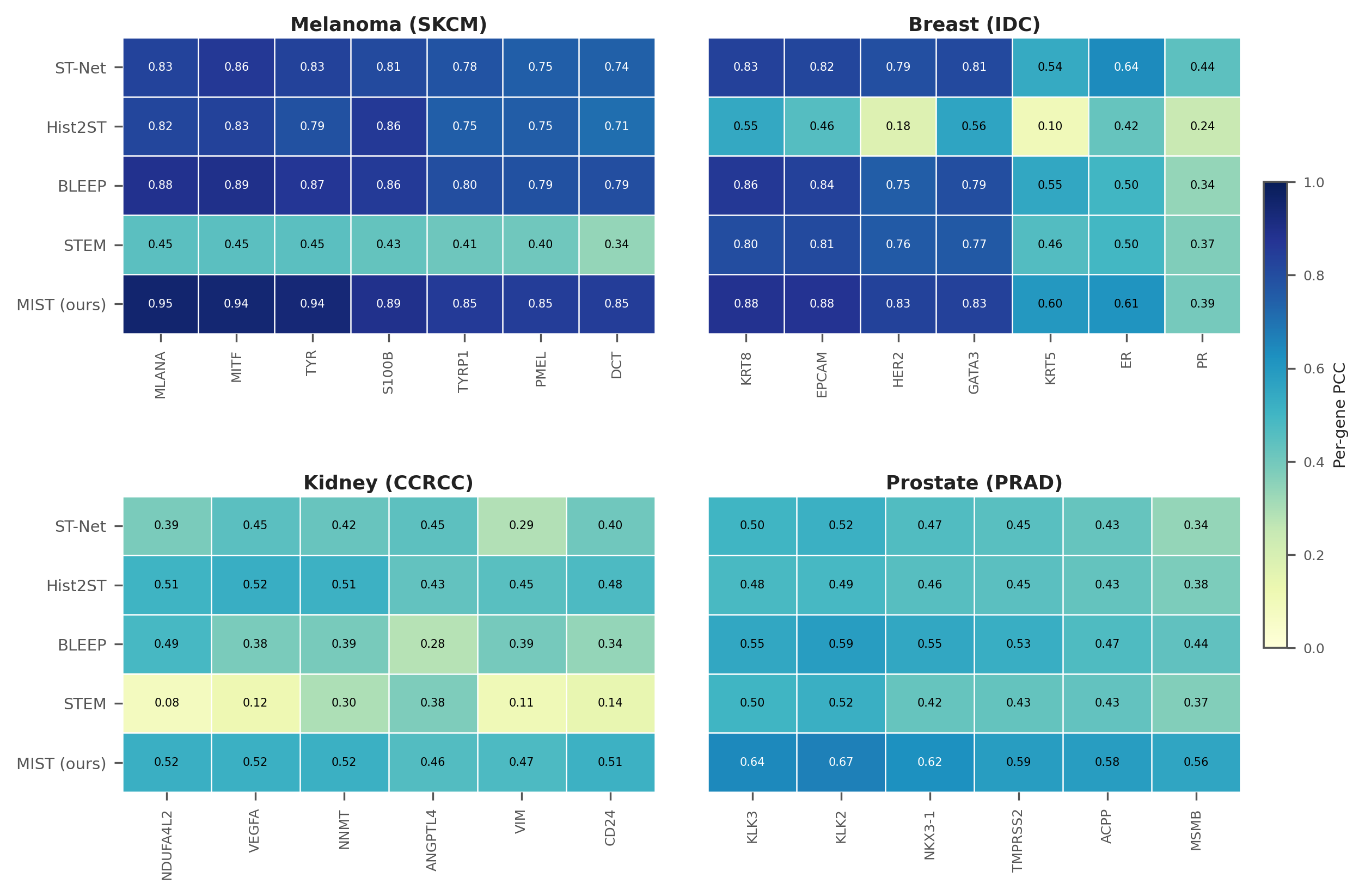}
\caption{\textbf{Per-gene recovery of canonical cancer markers across four cohorts.} Per-gene Pearson
on the 200-gene panel (melanoma/SKCM, breast/IDC, kidney/CCRCC, prostate/PRAD), mean over folds and
three seeds; darker is higher. STEM values for CCRCC and PRAD are computed from the folds available at
submission time.}
\label{fig:markers_supp}
\end{figure}

Figure~\ref{fig:markers_supp} shows per-gene marker heatmaps for all four cohorts whose canonical
markers are present in the 200-gene panel: melanoma (SKCM) and breast (IDC) from the main case study
(Figure~\ref{tab:casestudy}), together with kidney (CCRCC; clear-cell renal markers) and prostate
(PRAD). \morph{} attains the highest per-gene Pearson correlation on nearly every marker in every
cohort, most clearly for melanoma and the prostate panel (e.g., KLK3, KLK2, NKX3-1).

\section{Domain-Recovery Robustness to Cluster Count}
\label{app:krobust}
The spatial-domain analysis (RQ5) fixes the number of KMeans domains at $k{=}6$. Table~\ref{tab:krobust}
repeats it for $k\in\{4,6,8,10\}$ on the same $n{=}58$ held-out six-organ pooled slides
(StandardScaler features, mean over three seeds). \morph{} attains the highest mean ARI and NMI at
\emph{every} $k$, so its domain-recovery advantage is not an artifact of the chosen cluster count.
Absolute agreement is modest and the across-slide variance is large (median per-slide ARI ${\approx}0.04$),
consistent with the main text.

\begin{table}[H]
\centering
\caption{\textbf{Domain recovery vs.\ number of clusters $k$.} Mean ARI and NMI over $n{=}58$ held-out
slides (three seeds); best per column in \best{bold}.}
\label{tab:krobust}
\small
\setlength{\tabcolsep}{5pt}
\begin{tabular}{l cccc cccc}
\toprule
 & \multicolumn{4}{c}{ARI $\uparrow$} & \multicolumn{4}{c}{NMI $\uparrow$} \\
\cmidrule(lr){2-5}\cmidrule(lr){6-9}
Method & $k{=}4$ & $k{=}6$ & $k{=}8$ & $k{=}10$ & $k{=}4$ & $k{=}6$ & $k{=}8$ & $k{=}10$ \\
\midrule
ST-Net  & $0.070$ & $0.065$ & $0.055$ & $0.049$ & $0.085$ & $0.091$ & $0.092$ & $0.095$ \\
Hist2ST & $0.079$ & $0.073$ & $0.062$ & $0.055$ & $0.084$ & $0.087$ & $0.089$ & $0.091$ \\
BLEEP   & $0.090$ & $0.081$ & $0.075$ & $0.070$ & $0.088$ & $0.092$ & $0.095$ & $0.099$ \\
STEM    & $0.034$ & $0.031$ & $0.027$ & $0.025$ & $0.028$ & $0.028$ & $0.030$ & $0.032$ \\
\midrule
\morph{} (ours) & \best{$0.106$} & \best{$0.091$} & \best{$0.080$} & \best{$0.072$} & \best{$0.107$} & \best{$0.110$} & \best{$0.111$} & \best{$0.113$} \\
\bottomrule
\end{tabular}
\end{table}

\section{Computation Resources and Training}
\label{app:compute}
All experiments ran on a shared GPU cluster. Feature extraction and training used NVIDIA GPUs with
compute capability $\geq$~sm\_80 (A100 80GB, A100-PCIe, L40s, and H200); Volta (V100) GPUs are
incompatible with our CUDA-13 / PyTorch build and were excluded.
\begin{itemize}[nosep,leftmargin=1.4em]
\item \textbf{Encoders:} frozen UNI (ViT-L/16, 1024-d) and CONCH (512-d) patch features, concatenated
to 1536-d; no encoder fine-tuning.
\item \textbf{Model:} \morph{} has 4.88M trainable parameters and predicts
a whole slide in a single forward pass, with no iterative sampling.
\item \textbf{Training:} AdamW with a cosine schedule, 100 epochs with early stopping (patience~20),
nested validation, and a correlation-aware objective; three seeds per configuration.
\end{itemize}

\end{document}